# Evolution of Natural Language Processing Technology: Not Just Language Processing Towards General-Purpose AI

Evolution of natural language processing technology:
From "language" processing to general-purpose AI.
Recent trends from a Japanese point of view.

Masahiro Yamamoto

INFORMATION-TECHNOLOGY PROMOTION AGENCY, JAPAN

General Affairs Planning Department

Abstract

Since the invention of computers, communication through natural language (actual human language) has been a dream technology[1]. However, natural language is extremely difficult to mathematically formulate, making it difficult to realize as an algorithm without considering programming. While there have been numerous technological developments, one cannot say that any results allowing free utilization have been achieved thus far.

In the case of language learning in humans, for instance when learning one's mother tongue or foreign language, one must admit that this process is similar to the adage "practice makes perfect" in principle, even though the learning method is significant up to a point. Deep learning has played a central role in contemporary AI technology in recent years. When applied to natural language processing (NLP), this produced unprecedented results. Achievements exceeding the initial predictions have been reported from the results of learning vast amounts of textual data using deep learning. For instance, four arithmetic operations could be performed without explicit learning, thereby enabling the explanation of complex images and the generation of images from corresponding explanatory texts. It is an accurate example of the learner embodying the concept of "practice makes perfect" by using vast amounts of textual data.

This report provides a technological explanation of how cutting-edge NLP has made it possible to realize the "practice makes perfect" principle. Additionally, examples of how this can be applied to business are provided. We reported in June 2022 in Japanese on the NLP movement from late 2021 to early 2022[2]. We would like to summarize this as a memorandum since this is just the initial movement leading to the current large language models (LLMs).

Chapter 1 presents an overview of NLP in recent years. What is made possible through vast amount of textual data and parameters is explained through the example of GPT-3[3], which caused a major stir when it was released. In particular, the outline of Transformer, which is one of the key technologies, is being explicated to achieve intuitive understanding.

In Chapter 2, the potential applications of GPT-3 are presented to discuss the practical side of large-scale language models. Some of these services have already been launched, and while they have clearly demonstrated the usage of AI, they have realized unprecedented language-based exchange, thus setting them apart from existing services.

Chapter 3 discusses the ongoing studies in Japan in the field of NLP. What is the situation in Japan when the development of large-scale language model takes place mostly in English? The simple overview of this situation is discussed, along with the results of recent dialog competitions.

Chapter 4 discusses the report stating that a large-scale language model will become the fundamental base model of a general-purpose AI. As most human activities can be explained using language, a general-purpose language model describes most human activities within the model. Therefore, the possibilities of large-scale natural language models that can lead to general-purpose AI will be discussed a paper from Stanford University, USA. Moreover, the relationship between large-scale language models and robotics is discussed focusing on the perspective of direct interaction with the real world.

Chapter 5 discusses the developments, mostly international, in NLP technology beyond GPT-3. The author provides a snapshot of the current situation, which sees daily updates, including the construction of larger models and the method to improve performance while reducing the scale of the model.

Chapter 6 discusses the important challenges faced in the present NLP and natural language models. Language has a significant impact on society, and this fact will be re-examined.

Chapter 7 contains the summary and future prospects.

# Table of Contents

# 1  Introduction

In 2018, Google launched an AI assistant named Duplex [4] at the company's development conference, and performed a demonstration where it made a reservation at a given restaurant. Upon opening the Duplex application and using one's voice to specify the restaurant, number of people, date, and time, it will automatically book a reservation over the phone. As many restaurants do not offer online reservations, reservations can be made at any restaurant using Duplex as it can call the desired restaurant on our behalf.

Examples of actual interactions with the application sound natural. It also responds sufficiently naturally during phone calls to restaurants to make reservations. It did not cause any confusion in the restaurants where the reservations were made, meaning that they did not notice that Duplex was automatically conducting the whole exchange. In fact, the restaurant staff were surprised when they found out that they were not talking to an actual human.

When Duplex was launched in 2018, it required partial human assistance for 25% of its operations initially; the update report from 2020 revealed that 99% of its reservation operations were conducted through automatic response by the AI [5]. Moreover, Duplex had successfully made a million reservations by that point.

In addition, Google reported that it uses Duplex to automatically update business information on Google Maps and Google search [6]. Detailed information such as the opening hours of restaurants or whether the restaurant offers takeout service is updated automatically, without the need for manual updates.

Duplex not only looks up information necessary for making reservations through Google search but also updates the very information it uses by itself. Now, further functions have been added to realize additional applications, such as purchase movie tickets and reserve rental cars. When a person speaks to the application using natural language, the AI can perform all the tasks while offering unprecedentedly natural responses.

One of the key technologies that supports Duplex is the natural language processing (NLP). The general term for technologies that enable a system to simulate dialog through everyday language, and that enable the system to automatically perform a task and report the result of the same through natural language, is NLP. In fact, the Duplex example shows that NLP has various potential applications, ranging from interfaces to business logic.

Currently, cutting-edge NLP technology is capable of generating a natural sentence from a single word irrespective of automated processing and results from language inputs. One can even say that the replacement of even human creative work is becoming a realistic near-future prospect. Fig.1 shows an example (diagram) of NLP.

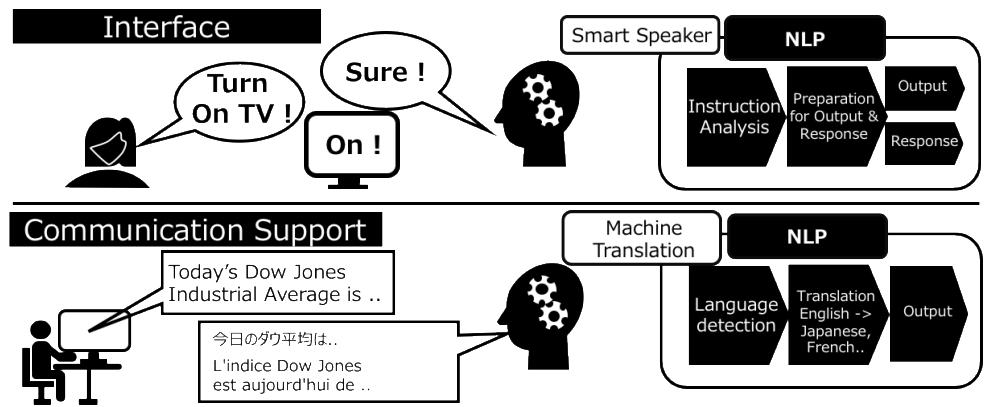

Fig. 1 Natural Language Processing with use cases

**1.1 Significant progress is supported by the rapid increase in the number of parameters**

One of the factors behind the expansion of Duplex applications is a new natural language model (NLM)[7] named bidirectional encoder representations from transformers (BERT)[8], which has large scale model parameters[9]. The rapid increase in the number of learning parameters is one of the reasons for the explosive development of NLP. Deep learning, which is the central technology for AI in recent years, characteristically improve the capability of AI through numerous learning parameters using unprecedentedly vast amounts of data.

The performance of NLP in the last few years has improved remarkably owing to the increase in the number of learning parameters. Fig.2 shows seminal NLP technologies at each historical moment and their parameter numbers at the time of their release. In fact, the number of parameters used by language models has increased by hundred to thousand-fold in last five years. Currently, the number of parameters used in Google T5+ is 1.6 trillion.

How do neural networks with this many parameters learn and optimize? How does one prepare the training data for such networks? These questions are important from intuitive and technological perspectives. For instance, GPT-3, which gained considerable interest due to its performance in 2020, uses textual data comprising four trillion words for its learning, and the required computational cost is estimated to be a few billion yen. Furthermore, it requires a vast amount of computational resources, estimated at computing on V100-class GPU cluster with 1420 GPUs for one year.

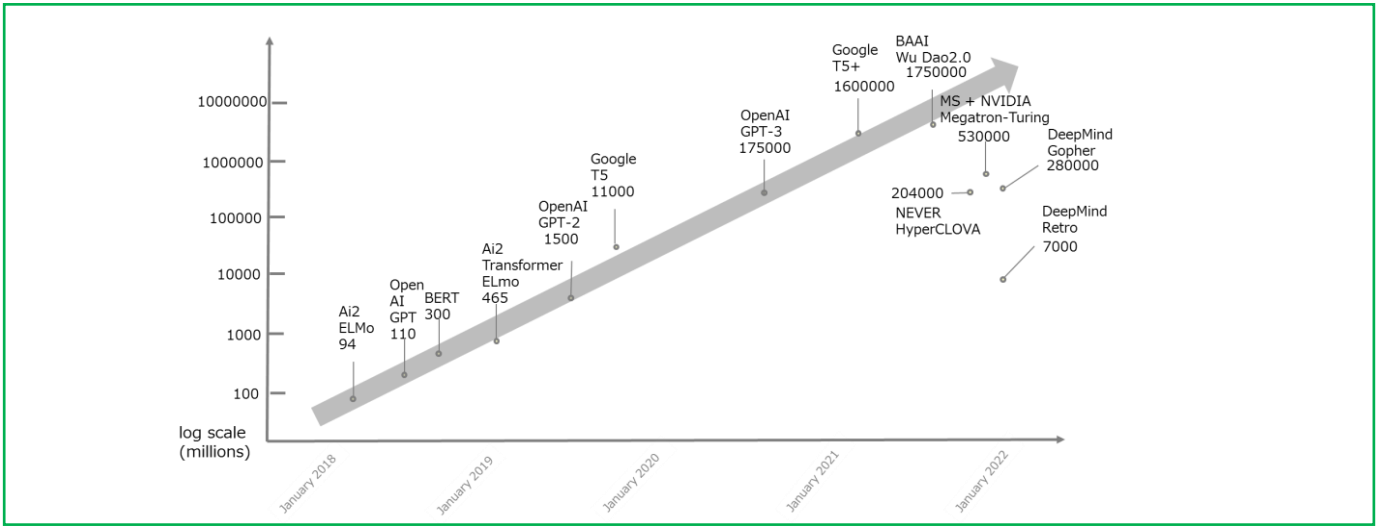

Fig.2 The parameter numbers in the language model[10]

Computational resources continue to evolve considerably with each passing year. For instance, GPU A100, introduced in 2021, is expected to have ~20 times more capability than V100 and enable learning with improved realistic times and computational scales. H100[11], which NVIDIA released in 2022, is said to be capable of building NLMs 30 times faster than A100. For instance, it can build a vast language model approximately the same size as GPT-3 in a week. Naturally, this makes it easier to enlarge the language model itself to ten or several hundred times larger than it is now. The computational capabilities will continue to increase remarkably as with the evolution of semiconductor chips. Thus, computational resources with high cost performance can be realized in the future, which make the facility planning crucial.

The NLMs released in late 2021 did not necessarily have numerous learning parameters[12]. This indicates that, through innovation, a model can have performance equal to models with a large number of learning parameters as long as it has the required number of learning parameters. This must be regarded as a new trend from the perspective of computational resources.

Interesting results have been obtained by increasing the number of parameters in GPT-3; a NLM successfully performed four arithmetic operations without explicitly learning its computing logic[13]. The aforementioned language model showed signs of performing two- and three-digit additions when the number of model parameters exceeded 6.7 and 13 billion, respectively. When the model parameters exceeded ~100 billion, it became capable of almost perfectly performing two- and three-digit addition and started to display signs of being able to perform two-digit multiplication and four-digit addition. As the number of parameters increase, the NLM acquires computational capability of addition, multiplication, subtraction, and division. Even though only textual data were used for training, a NLM acquired the function to perform calculations when the number of its parameters exceeded 10 billion. The vast amount of text data included the descriptions of calculations and a couple of corresponding examples. However, the capability of a NLM to learn calculations, without explicit training data for the same, is an astonishing result. Furthermore, it demonstrated the possibility that, only through learning via text data a NLM with a vast number of parameters is capable of learning the described procedure itself and integrating it as a function. Naturally, it also learns how to write programs.

**1.2 Breakthrough by Transformer that has enabled efficient round-robin**

The history of research and development of the NLP, which has been expected to be the main technological area in AI is long. Though there have been many efforts made since deep learning became the main focus of AI, no major breakthroughs were made immediately. This was partially due to the characteristics of natural language. Developments in deep learning include the so-called general-purpose computing on graphics processing units (GPGPU), which uses GPU[14] for purposes other than image rendering. GPGPU performs calculations using numerous (approximately a few thousand) product–sum operation units on the GPU in parallel.

However, in the NLP, it is necessary to deal with meanings, directives, pronouns, etc., while going back and forth in their context. Understanding texts, in particular, requires such processing. As such processing requires AI to remember specific parts while processing the whole in time series, it is generally not very compatible with parallel computing. Therefore recurrent neural network (RNN)[15], which is used to deal with time series, is commonly used in natural language. As the sequence-dependent relationship must be considered, it generates (wait for) adjustments between several dependent relationships, thereby complicating the parallel operations. Thus, GPGPU cannot be used effectively for the NLP involving sequence-dependent relationship. As the current mainstream in calculations using GPGPU prioritizes parallel and mechanical calculations, time-series calculations with dependency is intrinsically unsuitable for it.

To address this issue, the Attention[16] mechanism was developed. Using this mechanism, the connection of a word with its counterparts in a sentence is discovered by focusing only on the word. Note that the focus is only on these connections and not the grammar and mechanics. Fig.3 shows an overview of the attention mechanism. In the sentence "Mount Fuji is high, and the sea is beautiful," the attention is on "high." Then, it calculates how much weight should be placed on its connection with other elements in the sentence, namely "Mount Fuji," "and," "the sea," "is," and "beautiful." The figure shows this schematically by representing the connections through solid and dotted lines and the thickness of these lines. This operation is repeated for every element in the text. As the operation is performed on all the text that can be stored in the memory, in principle, the weight of the connection between elements that are far apart can be calculated.

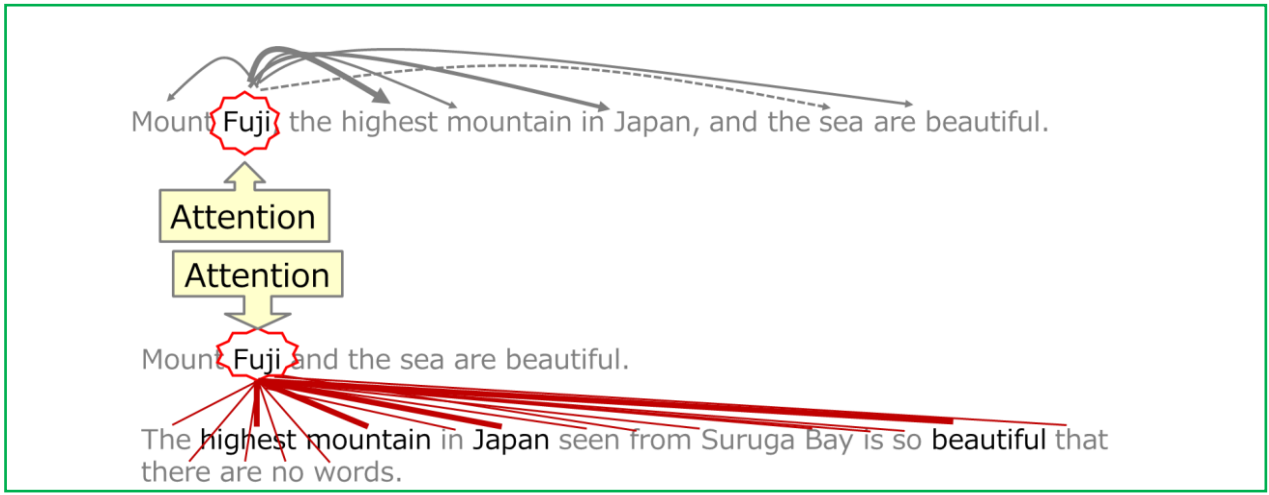

Fig.3 Schematic diagram of attention mechanism

As the text in the memory can be handled as though it is an image, parallel computation that compares the feature value(s) of one part of the text to the feature value(s) of other part becomes possible. In other words, even though NLP is intrinsically unsuitable for parallel computation, it can be processed through parallel computation using the attention mechanism. From a different viewpoint, one can also say that this reverts to the fully connected network as it calculates the connection between every element on the memory.

The effect of parallel processing becoming possible is enormous. Aside from the obvious fact that efficient parallel computation using GPGPU has become possible, common supervised learning[17] is no longer always necessary because, through full connection, the AI is only required to learn the weight relationships through text cross-referencing. Natural language learning cannot be easily evaluated, unlike human learning, so it is necessary to use existing texts for learning. Furthermore, some amount of learning data should be used as the training data for simulating tests within the context to teach questions and answers. The problem was that the cost of producing these pairs was extremely high. However, if it comes down to what is known as a fill-in-the-blank question, so it can turn, in principle, any text into training data with almost zero cost. BERT mostly employs two simple learning methods namely masked language modeling (MLM), which performs what is known as fill-in-the-blank, and next sentence prediction (NSP), which evaluates the similarity of sentences.

Fig.4 shows the concept of MLM. As attention is based on bidirectional referencing, one can say that the biderectional referencing itself is the learning. If normal text data is present, one can simply mask a part of it and turn it into training data with the correct answer. This procedure, where data itself is changed/worked on to give the correct label to the original data, is called self-supervised learning. As it does not explicitly supply training data, it is attracting attention as a typical example of self-supervised learning. It has been demonstrated that masked estimation questions, where parts of a sentence are masked, which is common in tests, is a valid framework for self-supervised learning, and significant progress has been made using this approach.

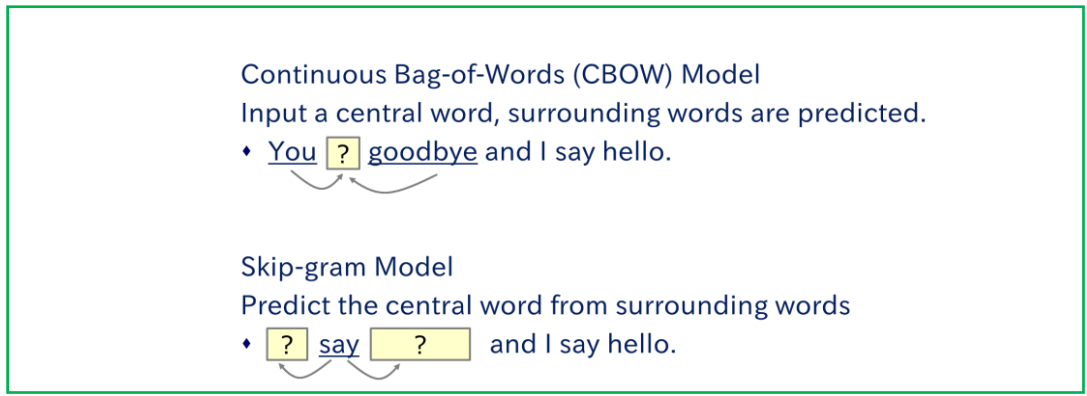

Fig. 4 Masked Language Model (MLM)

NSP compares two sentences and judges whether the second one is an appropriate sentence to follow the first one (Fig.5). This can be easily determined from cross-referencing information. The knowledge regarding training data verified in BERT showed that normal text data can also become high-quality training data[18]. In other words, a large amount of training data will be spontaneously generated without modifying the text data. Learning by MLM and NSP, NLMs successfully learned relationships and texts simultaneously. Furthermore, it learned calculation functions.

| Sentence Pair | Label |
| --- | --- |
| My mother cooked pasta. It was delicious. | isNext |
| A whale was swimming. He solved a math problem. | NotNext |

Fig. 5 Next Sentence Prediction (NSP)

The center of this attention-based basic machine learning is the technology called Transformer. Transformer enabled usage of normal texts as training data, as well as processing of numerous parameters by efficiently using calculation resources in parallel. Transformer, a technology that has attracted attention toward BERT, has made it possible to efficiently learn a vast number of learning parameters and express the corresponding functions.

**1.3 Features of natural language processing by Transformer**

Let us introduce additional keywords necessary for understanding Transformer. These keywords, Parameter scale, Natural language model, Zero(Few)-Shot learning, Language benchmark, and Multilingual support, have become more than simply the characteristics of contemporary NLP technology[19].

Parameter scale

For instance, when there are more parameters, useful results can be expected by expanding the reference area during learning and its reflection on the model. Thus, when the parameter scale is enlarged, the limit of the existing models will also be pushed up, and their performances are also expected to improve. Indeed, performance assurance through parameter scale enlargement in Transformer has been verified[20], indicating that the enlargement of parameter scale directly leads to performance improvement. In principle, the larger the parameter number, the better the performance of the NLM. Therefore, larger parameter scales are currently being tested. Meanwhile, methods to compress the enlarged model to make it efficient are being investigated

Natural language model

Generally, a NLM is the core of machine learning for NLP, and it is the foundation of NLP. One element of NLP is to determine the position of a given sentence in a sentence assess. In Transformer, the model is built based on the probability of appearance of a given word at a specific position in a sentence is determined through cross referencing preceding and succeeding words.

NLMs comprising several parameters and following large-scale prior learning[21] through vast text data acquired a surprising amount of knowledge from their training data. This knowledge was mainly acquired by learning relationships among words. These models not only acquired knowledge but also learned the logical connections between descriptions, which cannot be acquired only by learning relationships among words. Therefore, they not only raised the quality in text generation tasks, such as text summarization or conversational response that is accurate and compositionally correct but also became capable of performing functions other than language tasks, such as calculation or automatic programming from specifications, through a single language model.

Zero(Few)-Shot learning[22]

The prior learning model contains all referential relationships described in the text of the training data. Therefore, domain tuning is possible without an example or with only one or two examples, thus enabling wide-range output of arbitrary tasks.

Few-shot learning is when a small number (10 to 100) of examples are offered for the task (during inference), and one-shot learning is when only one example is offered. Similarly, no examples are offered in zero-shot learning. Zero and Few-Shot learnings are highly dependent on the performance of the prior learning model. However, today, almost the best performances according to the benchmarks of successful task completion have been achieved without any task examples or only with one task example without learning each domain, which was previously necessary.

GPT-3, in particular, uses prompt learning[23]. Due to its high performance, it has become an almost *de facto* tuning method for subsequent NLMs. In prompt learning, simple exchanges are tuned on the dialog UI screen by presenting examples of simple sentence pairs according to the type of shot learning. This is commonly known as sequential learning. In this case, additional learning methods can be understood as the process of additional adjustment to determine which relationships among the vast number of text pairs in the NLM the domain should prioritize.

Language benchmark

The evaluation of NLM is extremely important when using the model. For instance, a person can determine the naturalness of a simplesentence. However, quantifying this naturalness in number is fundamentally difficult. If quantification of evaluation was at all possible, NLP can be conducted simply by supplying the quantified evaluation to an algorithm. With this situation in the background, several benchmarks[24] have been devised as evaluation indexes for NLP based on training data written by humans to enable objective evaluation through tests. For instance, they could include questions and answers, text generation, natural language inference of logical connections across several stages, solutions through referencing, and the resolution of ambiguity in the meaning of a word and its translation ( common in tests with the Japanese language).

The time between the proposal of a benchmark and for an AI to exceed the human scores in it is becoming decreasing every year[25]. When examining several examples, including the widely used MNIST[26] and recent GLUE[27], it shows that AI took more than 15 years to exceed human scores in MNIST, which was proposed before 2000, it only took less than 3 years in GLUE, despite the fact that it was proposed recently and the contents of benchmarks became advanced.

Since deep learning became widespread in the 2010s, the score is improving rapidly, and despite the fact that GLUE was proposed in accordance with this era of deep learning, AI score exceeding human scores was reported only after two years. Earlier, research and development in NLP system built systems only to improve benchmark and achieve higher scores. Today, however, this approach has changed where more general NLP systems are built, which are then examined through various benchmarks. In other words, it has shifted from learning for the sake of tests to conducting general and wide-ranging learning before taking tests. This indicates the fact that technologically, the performances of the current general-purpose NLP systems not specialized in any individual tasks and domains are equal to those of the specialized systems.

Currently, existing benchmarks are becoming useless owing to the development in NLP technology in recent years, and new benchmarks are being proposed[28]. Additionally, benchmarks are being used for assessing performance and conducting detailed diagnoses of operations. Moreover, most benchmarks focus on specific task or domain because creation of universal benchmark is difficult.

However, several problems can arise while realizing applications of LLMs when these applications are used in society. Therefore, researchers have been conducted to determine new benchmarks that improve the shortcomings of existing benchmarks and are more suitable for social implementation to further refine NLP systems. Simultaneously, other aspects are expected of new benchmarks. The situation demands a serious examination of the risks of natural language models through the benchmarks.

Many people believe that the current benchmarks are insufficient for the evaluation of several important risks. For instance, let us consider a situation where the output provided by an application that uses a natural language model for a given input is incorrect. Can we be sure that people will not believe this information to be true? For instance, if this application usually offers information that is almost 100% trustworthy, when as an exception the information it has given is incorrect, people would normally continue to trust the application. This would then become the trust bias and people would believe the incorrect information. Under such a situation, problems exist even when there is no malicious intent. Thus, usage of natural language models inherently contains significant risks.

At this point, the only solution is to avoid evaluating natural language models only through benchmark tools and let people actually test and verify them. Thus, it is important to conduct further discussions on risk reduction. As discussed earlier, the reproduction of harmful social stereotypes by natural language models are becoming a major problem.[29] In fact, one can say that studies on such problems are only at their initial stage. There are many people who are hoping for further improvements in benchmarks for better evaluation of risks.

Multilingual support

Diversity in a language is also an extremely important aspect of natural language models. The structure of a language is usually of an extremely localized nature, as it is heavily influenced by the place and culture in which it was born. A pretraining model is expected not only to learn cross-referencing in one language but also to learn the universal relationships shared across multiple languages. Therefore, the learned universality across multiple languages is essentially the distributed representation that should be contained within a natural language model. In other words, it is a feature representation vector that should acquire. If distributed representation has acquired universality, it means it has acquired various relationships corresponding with objects and circumstances at a more fundamental language level. Thus, one can infer that even when language is different in each country, it does not cause problems for language models. For this reason, by using language data that is as wide-ranging as possible to build a natural language model, it is possible not only to achieve two-directional multilingual machine translation but also to make the natural language model itself of higher quality.

In principle, bidirectional text data of the translation source and language to be translated into is necessary for translation. However, when a natural language model acquires distributed representation, machine translation of a language is expected to become possible, even when no sufficient bidirectional text data for its learning is available. In fact, Google announced during its development meeting on May 2022 that a single language approach, whereby a language model learns translation of a new language even when there is an insufficient amount of training data, is possible, and it newly added 24 languages to its translation service[30]. While these 24 languages are not regarded as major languages, the total of their users are as large as 300 million people. Thus, there is high expectation for the expansion this next generation translation will offer.

As there are more than 7,000 languages in the world[31] (and strictly speaking, distinguishing among these languages is considerably difficult), only a small part of these languages can be translated today. It is known that the languages around the world are derived from the top language families[31]. While only a small number of languages have been modeled thus far, these have still had impacts on the improvement of distributed representation. As the number of modeled languages will increase in the future, one can expect that LLMs will acquire further universality. Moreover, accommodating more languages is known to be useful for responding to unknown languages. Machine translation accommodating every language in the world is not only an important technological challenge in itself, but also an intellectual challenge to understanding the language itself, considering the current position of natural language models.

These keywords will help us to understand NLP by Transformer. If a natural language model possesses all these characteristics, one can say that the expression "practice makes one perfect" applies to AI as well. As the results of current NLP can appear to understand the meaning of language, the phrase 'natural language understanding' is beginning to be used. This is because the system appears to understand the content of a text and processes it considering its meaning. While whether the system truly understands the meaning is still being debated, it is undeniable that extremely advance processing has become possible.

## 2. Natural language processing technology application

GPT-3 was released on July 2020 as the third large-scale natural language model by OpenAI as a result of NLP. Its impact grew larger with time[32], and now it is highly possible that, in a few years, it will be regarded as the turning point of AI itself. GPT-3 is a natural language model with the Transformer that has 175 billion parameters trained with several trillion words.

GPT-3 has extremely advanced NLP capability, and it processes and generates text as though it is a human. For instance, when a user inputs text on the prompt window, where text inputs and outputs are fed, GPT-3 returns an output in the form of an email, tweet, or trivia quiz with consistent topics according to the purpose. Thus, an environment where, in principle, the entire text of an email, exchange with customers, social media entry, and news article, or at least parts of them, can be automatically generated suddenly emerged.

The usage policy of GPT-3 by OpenAI supports commercial application, and in response, business application started instantly. Using such a natural language model, a new start-up can launch an application; thus, GPT-3 and its equivalent models can deliver easy access to the enormous power of AI to the people who are investigating the application of NLP around the world[33]. Indeed, when the author of this paper searched GPT-3 as an author in Amazon[34], he found publications where GPT-3 is listed as the co-author. While it appears that human help is still needed to fine-tune the texts from GPT-3, the fact that these publications are for sale indicates an extremely important turning point.

**2.1 Examples of business applications of GPT-3**

For example, OthersideAI[35] can generate an entire email from only the keywords of the main text or from its opening line. Examples of such text generation from keywords also include Broca[36] and Snazzy[37] used for writing advertisement copies or campaign contents. Both companies offer trials. Thus, these innovations in automatic text generation can be experienced immediately. These examples are thoroughly discussed in the subsequent paragraphs.

While email has occupied a significant position as an important communication tool since the early days of the Internet until today, in recent years it has also been regarded as a representative of decreasing productivity, or indeed its cause. OthersideAI solved this problem by overcoming the productivity loss caused by writing e-mails, and instead is utilizing GPT-3 to increase productivity. In principle, let us think of a combination of a question and an answer. Not only is this generated each time, but it also enables detailed adjustment for each scenario even if they have similar contents, allowing for minor changes and the addition of necessary information for each scenario. Thus, it can accommodate several scenarios and situations, such as sales, support, and marketing, that conventionally required reorganization each time, without any problem as a common engine. Their demo offers an example of an email where a complete text was generated only after input of keywords. It shows that it only requires entry of necessary and sufficient words, and the remaining email is generated automatically.

Seen from another perspective, this automatic text generation function can also be used as typing assistance. When writing e-mails, using one's own expression for every word and clause is difficult to begin with, and for business correspondence, a large portion of one's email is standard expressions. In any case, there is no problem with the text itself if most of it was automatically generated. This means, for instance, when writing a text through eye gaze input, text is generated only through the input of initial words and the rest of the process only requires only checking and minor corrections. The benefits of such a process are in immeasurable.

Such text generation through input assistance was not expected at the start of the service. Instead, its possibility was recognized during the verification and analysis of automatically generated texts, which was then chosen as an area of its application. Indeed, one can say that it is one of the easiest input assistance tools developed so far.

Broca and Snazzy that offers advertisement service fully utilizes text generation through keyword input. The fact that a campaign contents that takes context into account can be automatically generated from the input of one word, or from extremely short but impactful advertisement copy, appears to be a highly convenient service.

With this automatic generation contents by an AI, Broca is attempting to expand contents marketing. Its strength is the fact that contents with high quality that used to take a few hours, or even a few days, to prepare can be generated in a few minutes. To quell doubt on its quality and guarantee its good performance, all the contents in the website of Broca is being generated through the service it offers. Naturally, this indicates their confidence in the quality of the contents their service offers, which can provide everything from contents for advertisement to social media posts.

For instance, its website lists an example of generating campaign contents for a new product. It is a three-step process. In reality, the first two to three sentences entered at the first step are the only input required from the user. The content is automatically generated only after this input, and one can obtain promotion content of any type consistent with its context.

Several examples of BERT as a basic element have been reported in the literature. FORETHOUGHT[38] provides a question-and-answer search AI agent called Agatha. Agatha uses knowledge-based articles and help desk templates to answer general questions through e-mails and Web widgets. It offers immediate answers to users. One feature of Agatha is continuous training through machine learning and NLP for the purpose of constant improvement with time. It can already accommodate 100 languages and is offering language solutions globally.

Gong[39] provides a next-generation CRM that performs speech analysis of conversations with customers. As a result, several thousand sales teams have completed business using the NLP function of Gong, shortening their selling cycles. As 99% of the information one shares with the customers does not reach CRM and the remaining 1% is heavily filtered, they are not useful for sales. For this reason, Gong extracts elements with higher order insights that are relevant for information necessary for real business dealings by applying NLP on information from the customers. It copies e-mails, phone calls and video calls from customers, then analyzes all the elements that can potentially lead to insights on whether the clients are prepared to propose updates on products or whether there are risks of losing business by utilizing machine learning.

Moveworks[40] provides a bot that can autonomously solve various problems of the staff when they are using IT. In addition, it is enabled to understand conversation and ambiguous questions through its NLP function. Starting in the summer of 2018 where it autonomously solved 20% of IT-related problems for a customer without human support, the bot can currently solve up to 40% of these problems, and under certain circumstances, it can solve 65% of problems. In other words, 65% of problems that arise when using IT can be autonomously solved. See below for examples.

Example   *'Question' → 'Automatic answer'*

*'I want to edit a PDF file.' → 'I will issue an Acrobat Pro license.'*

*'I want a new keyboard because I'm working remotely.' → 'Which one of these keyboards do you want? Please select the number.'*

*'I forgot my password.' → 'Please follow this link to create a new password.'*

Application of this function is clearly not limited to the IT field; this AI function has expanded into the fields of human resources, general affairs, and finance.

OBSERVE.AI[41] analyzes the work of contact centers to optimize the same. Based on the experience that, at a contact center, only 1%, 2%, or even fewer phone calls will reach the other person, OBSERVE.AI applied speech analysis to this unexplored field. By using transcription of speech to text and NLP, they can discover the points of interest within a conversation, and they successfully used it to improve the customer experience. To better understand the mood of customers, they apply supervised learning to conversations at call centers, which led to better customer experiences.

**2.2 Technologies that apply GPT-3**

**DALL・E for flexible image generation through language and CLIP for flexible caption writing.**

OpenAI released DALL・E[42], which is capable of generating images from text. It learns 12 billion parameter versions of GPT-3 through text and image pairs. It reads the contents of the texts entered and can generate images that do not exist in its learning data through combination within the model. The released examples[43] show that it can generate an image (to be precise, an image group) from descriptions that are seemingly complete when a person reads it, but it becomes clear that detailed conditions are not given when one actually tries to draw from them, such as "avocado-like chair" or "a vase on the stool by the bed."

In the training of Transformer, elements that are seemingly unrelated are reflected in the training by weighting their necessary relationships under every condition. Thus, one can say that it is extracting all the necessary conditions for actually generating images, i.e. the detailed conditions necessary for drawing that are obvious conditions and can be obtained by pretraining the already learned commonsense, for instance light direction, shadow, texture, and the direction of the bedside. Thus, the missing conditions are inferred to be necessary and sufficient conditions, enabling it to make a drawing that is a resolved picture without any problem. These conditions include various transformations, such as personification and the usage of emoji. It also appears to be capable of considerably flexible transformation, within the range possible from the interpretation of the text, such as simplification or the display of details. As it includes the information complemented during image generation and, in a sense, re-links the complemented texts and images, this suggests another usage as an advanced automatic caption generation function, and examples of such usage are available.

Contrastive Language–Image Pretraining (CLIP)[44] with the announcement of the DALL・E performs the high-precision prediction of image and text snippet (caption) pairs through the new type of pretraining that combines an image encoder[45] and text encoder. There are still some major problems in image and its text description. The goal is to achieve what is known as the zero-shot category, meaning that high-precision text caption can be predicted for image data not used for its training.

One of the key ideas is a method that is simple but effective, namely the learning through a large amount of data, which has been shown in many examples thus far. While ImageNet uses 14 million images for its training, CLIP uses a large-scale dataset that contains 400 million pairs. Moreover, CLIP training uses pairs of various images and their captions on the Internet to realize the zero-shot function. In the actual training, an image and its corresponding caption are not directly optimized. Instead, it predicts the text label that has the closest relationship with the image.

CLIP selects a text caption from 32,768 randomly sampled captions. This is efficiently performed through Transformer. Let us look at an easy example. If the candidates for the caption include "a dog running," "a cat sleeping on a sofa," "a playing dog," and "a dog sleeping in its kennel," and the image shows a "cat," "a cat sleeping on a sofa" has the highest probability to be the correct caption and is therefore selected. Images and text captions are linked through repetition of this process. Distributed representation of images and distributed representation of texts are being learned. Languages and images are strongly linked through the connections at the word and image fragment level. By using CLIP, labeling now done by humans can be completely automated. As labeling of images is very expensive, there are high hopes for its application.

**Power Apps[46]**

Microsoft is in partnership with OpenAI. Afterthe release of GPT-3, it signed an exclusive contract with OpenAI for its usage[47]. The development platform of OpenAI has moved to Microsoft Azure, its cloud environment. During the developer's meeting, Build2021, it was announced that this powerful natural language model GPT-3 will be used for Power Apps, and several demonstrations were conducted. Power Apps, the released product, is popular as a low-code application development platform among wide-ranging users, from so-called Sunday programmers to professional developers whose job is programming.

When GPT-3 was integrated into Power Apps, it became possible to build applications from descriptions in natural language without considering codes and mathematical formulas. It automatically converts explanations in English into Power Fx. As the NLP function of GPT-3 is integrated into Power Fx[48], it also became possible to build applications by supplying samples through prompts.

The new NLP function enabled by GPT-3 will be integrated into famous major products such as Azure in the future[49]. As the exact same result as the description through mathematical formula can be achieved through simple words, anyone can create an application that even includes AI.

Thus, Transformer and Attention, which is one of its core technologies, conduct learning with an extremely large number of parameters containing vast amounts of data using deep learning, which is the basic technology of AI software today, as well as GPGPU, which enables efficient usage of GPU as hardware. This situation was explained through the example of GPT-3, which caused quite a shock when it was released. The technology improved even further following this, as it generated competition to release new technologies. Some of these will be discussed later.

## 3 Trends in Japan
### 3.1 Accommodating the Japanese language

It would be difficult for a single Japanese company to create a super-large-scale natural language model as good as the GPT-3. Thus, we must consider how NLP technology should be approached in Japan. For instance, one approach is to have several organizations collaborate to build a super-large-scale natural language model. If one is to build a natural language model on the scale of GPT-3, the necessary calculation resources are estimated to be equivalent to using RAIDEN of RIKEN for 24 hours every day for a year. (As mentioned earlier, if limited to natural language, the calculation resource capability improved by 600 times in the last four years. This speed of improvement is projected to stay the same according to the roadmaps of various companies involved in GPU. Therefore, one can also take a position that, as long as hardware can be updated, the calculation resources per unit time will eventually stop being a problem.) However, aside from the calculation resources, there is the issue of how to collect training data. This is a problem caused by the fact that we use Japanese, which is a language unique to the country of Japan. Even though, as long as there are texts, they can be used as training data, a significant amount of text data that is comprehensive and without bias is necessary to build a practical general-purpose natural language model. This is an especially serious issue.

Another approach to address this issue is to improve the algorithm. For example, the same level of capability to that of GPT-3 was observed in PET/iPET[50] only with 230 million parameters. Moreover, it exceeded the score of GPT-3 in SuperGLUE[51], which is a benchmark. It is a semi-supervised learning method where additional training data is generated from few-shot examples using the preleant ALBERT[52]. PET functions by first converting entered examples into cloze-style phrases. These are used to fine-tune natural language model ensembles. Then they are used to attach annotations to large unlabeled datasets to generate datasets with soft labels. The final model is fine-tuned through the soft-labeled data. This soft-labeled training data is used for the main training. In this case, the training cost, including that of resources, can be reduced to about 1/500, meaning that if the computing environment were the same, what took 500 days before would be ready in less than one day of calculation. It enables one to develop new strategic methodologies, such as shortening of training time or scaling down of the computing environment.

### 3.2 Resources in Japan

For reference, the representative natural language models (BERT) that are available in Japan are listed below. Most of them take the form of free use open source licenses[53], and one can use them at one's own company. As it provides a general-purpose model that already completed pretraining, it enables trials to easily be run on tasks because one does not need to build pretraining from scratch, even though it will require zero (few)-shot learning when used at one's own company. Currently, there are many examples of business applications using BERT, which is relatively agile, instead of using super-large-scale GPT-3. Moreover, it is also true that there is still a large scope for innovation as the underlying technology, as the example of PET/iPET shows. It is important that today, by using the BERT pretraining model, each user actually can apply it to problems to be solved and tasks to be performed. The BERT-type Japanese model (Large) released by the Kawahara Lab at Waseda University is one of the most complete models in Japan[54].

Table 1 Japanese BERT Models (Sources: Each URL)

| Producer | Software framework | Used Data | open source license | URL |
|---|---|---|---|---|
| Google | TensorFlow 2 | Japanese Wikipedia, etc. | Apache 2.0 | https://github.com/google-research/bert/blob/master/multilingual.md |
| Kyoto univ. Kurohashi-Chu-Murakami Lab. Wasade univ. Kawahara Lab. | TensorFlow 2 PyTorch Transformers | Japanese Wikipedia, CC-100 | Apache 2.0 (Bert base) CC BY-SA 4.0 (RoBert base) | https://nlp.ist.i.kyoto-u.ac.jp/index.php?ku_bert_japanese https://huggingface.co/nlp-waseda/roberta-base-japanese https://huggingface.co/nlp-waseda/roberta-large-japanese (Newer RoBERTa-base) |

| Tohoku Univ. Inui-Suzuki Lab. | TensorFlow 2 PyTorch Transformers | Japanese Wikipedia | Apache 2.0 | https://github.com/cl-tohoku/bert-japanese |
| NICT | TensorFlow 2 PyTorch Transformers | Japanese Wikipedia | CC BY 4.0 | https://alaginrc.nict.go.jp/nict-bert/index.html |

GPT Model and CLIP

| rinna corp. (GPT) | PyTorch Transformers | Japanese C4, CC-100, Wikipedia | MIT | https://huggingface.co/rinna/japanese-gpt-1b |
| rinna corp (CLIP) | PyTorch Transformers | CC-100, Wikipedia, CC12M | Apache 2.0 | https://huggingface.co/rinna/japanese-clip-vit-b-16 |
| rinna corp (CLOOB) | PyTorch Transformers | CC-100, Wikipedia, CC12M | Apache 2.0 | https://huggingface.co/rinna/japanese-cloob-vit-b-16 |

On January 2022, the Japanese GPT language model was launched[55]. Rinna, which became independent from Microsoft, released a natural language model specialized for Japanese with 1.3 billion parameters. It is provided under the MIT license and can be used commercially. The release of a Japanese GPT model that is more advanced than BERT is attracting attention, as it can be easily used in a relatively short amount of time after only few-shot learning, such as prompt learning or fine-tuning according to the purpose of each practical application. On May 12, 2022, the same company released CLIP and its improved version, CLOOB[56]. The licensing of both of these allow for commercial usage. They will be discussed together. For more information on recent Japanese LLM, please refer to the summary site[57] and the LLM study group[58] set up by top Japanese researchers.

**3.3 Discussions within Japan**

The LINE dialog system won two prizes during the dialog system symposium competition of The Japanese Society for Artificial Intelligence, and in its evaluation, it was commended for being indistinguishable from humans. LINE entered the competition with a dialog system utilizing the large-scale Japanese language model[59] developed with NEVER from South Korea. LINE won first place at the Open track and Situation track of live competition 4 at the dialog system symposium. "A HyperCLOVA[60]-based Chatbot with Unified Persona-Consistency Consideration and Knowledge Base" won top scores in two categories.

It is inferred that development took about a year since the release of the general-purpose natural language model by LINE in November 2020. At the point of its release, it was the first super-large natural model in Japanese, and it utilized a super computer with performance capability exceeding 700PFLOPS. It has more than 175 billion parameters, and more than 10 billion pages of Japanese data was prepared. Indeed, looking at the actual development history, up to 39 billion parameter models can be used in its Japanese model, and up to 82 billion in can be used in its multilingual model that includes Japanese.

During the explanation for developers in November 2021 by LINE, a detailed evaluation of the following topics was offered: Its natural response, i.e., whether conversations are smoothly conducted; if it is following a topic, i.e., whether it can follow the topic in its responses; if it is providing a topic or asking a question, i.e., whether it can provide a topic at each response; and its achievement of goals, i.e., whether the purpose of the dialogs are being achieved. According to the explanation by LINE, the result of performing these evaluations on the natural language models with different parameters during various tasks showed that the 39 billion parameters model scored the highest at almost every task. However, there were also aspects that require further improvement. For instance, the response to users with negative feelings requires a considerable amount of improvement.

It is an important result indicating that the preparation and application of large-scale natural language models of Japanese have started, and that they can achieve satisfactory results. In 2023 July 4th, NICT reports that it has created a pre-training LLM with high-quality Japanese text data.[61]

## 4  Large-scale language model that can potentially lead to general-purpose AI
### 4.1 Foundation Models

At Stanford University, very large natural language models are being proposed as foundation models. This is because the result of the evaluation of multimodal activities using them demonstrated that they can be the basic model not only for language by also for various AI tasks[62]. At their workshop[63], they discussed not only the technological aspects, but also much wider subjects, taking into account the fact that the realization of general-purpose AI is becoming realistic, which they believe will play an important role in the future society.

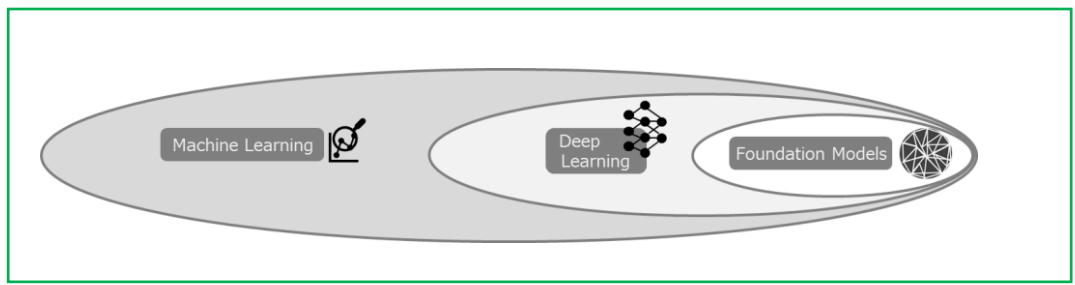

Fig. 6 Foundation Models is trends of Deep Learning[54]

As a basic concept, Deep Learning has been developed around features, while it is said that foundation models are further developing in this vein and placing an emphasis on "function." Fig.6 shows the overview of AI trends.

The large-scale natural language models that were trained with large-scale and wide-ranging data, in other words the foundation models, are models that can accommodate diverse downstream tasks[64] (BERT, GPT-3, DALL-E, CLIP, etc.), and they clearly have achieved a paradigm shift. Naturally, foundation models are extensions of conventional deep learning, and even though they are based on self-supervised learning and transfer learning, they do not appear to be particularly unique at first, aside from their vast numbers of training data and training parameters. However, their performances appear to offer new emergent functions, and their effectiveness has been demonstrated in an extremely high number of tasks.

Stanford's argument starts at this point. One model can accommodate every task, even multimodal inter-task tasks, providing powerful leverage, but it also means all its downstream adaptive models inherit the defects in the foundation model. Thus, one must pay the utmost attention to its influence. With foundation models, which have a far larger number of parameters compared to preceding deep learning, this is even more pronounced. Despite wide-ranging benchmark results, there is still no sufficient understanding of how the models might function, how they can potentially fail, and what is most important for emergent characteristics.

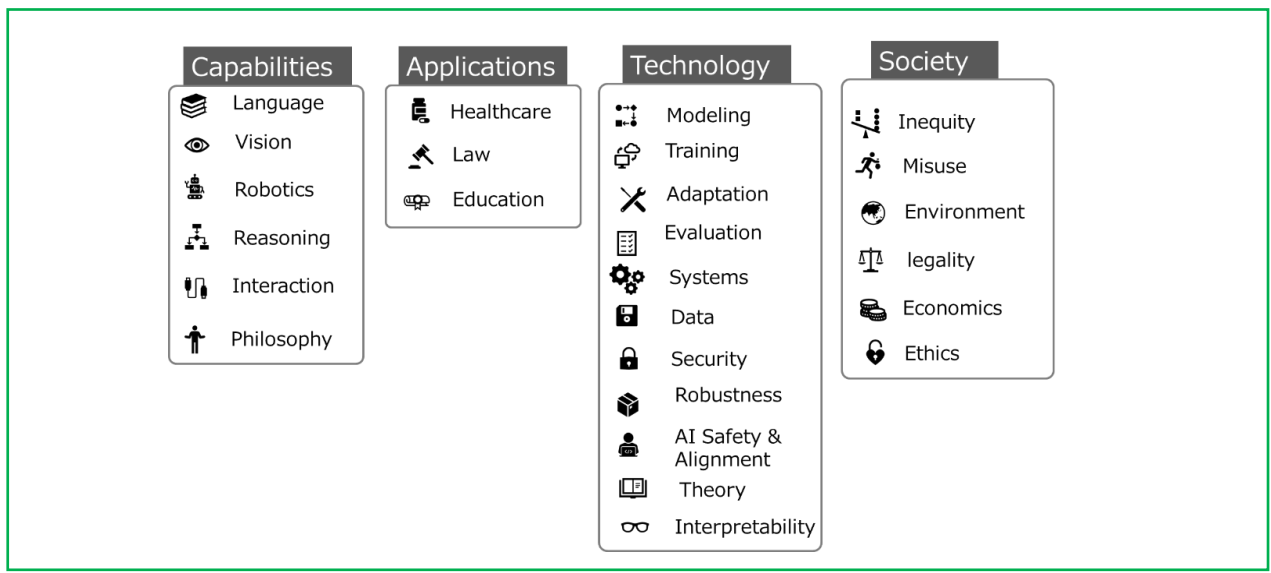

Fig. 7 The scope of discussion in Emergence and homogenization[54]

One of the guiding principles of Stanford today is to take this situation seriously and to systematically address it while following the major flow in the field. They examine the potentials and risks of various subjects, from the functions of foundation models (language, vision, robotics, reasoning, interaction with human, etc.) to their technological principles, (model architecture, learning procedure, data, system, security, etc.) and they discuss their possible applications (e.g. law, healthcare, and education) as well as their possible social impacts (e.g. equality (inequality), misuse, financial and environmental impact, and legal/ethical considerations). A diverse number of topics have been/should be discussed. Fig.7 lists the important keywords. In addition to technological aspects, they also seriously consider how technology groups are positioned within society, clearly indicating the possibility and influence of these technologies.

**4.2 Essence of natural language models**

Now, why do natural language models have such multimodal character? This was also debated during the workshop. In principle, a consensus is forming that this is due to an extremely simple reason. Namely, as language has the capacity to represent almost all human activities, converting it into objects and circumstances is thought to be relatively easy once its representational ability is modeled. Therefore, as natural language models learn distributed representation without any distinction, its training data is diverse and wide-ranging. Thus, it would contain a large amount of objects and circumstances, allowing it to reach the level where it is capable of text description regardless of modality. Lastly, when this distributed representation can be converted back to objects and circumstances, they are recreated as multimodality.

Fig.8 shows a comparison of a common natural language model and foundation models. One can understand foundation models as something that developed from natural language models, which only handle texts, and expanded into multimodality, while still centered on text.

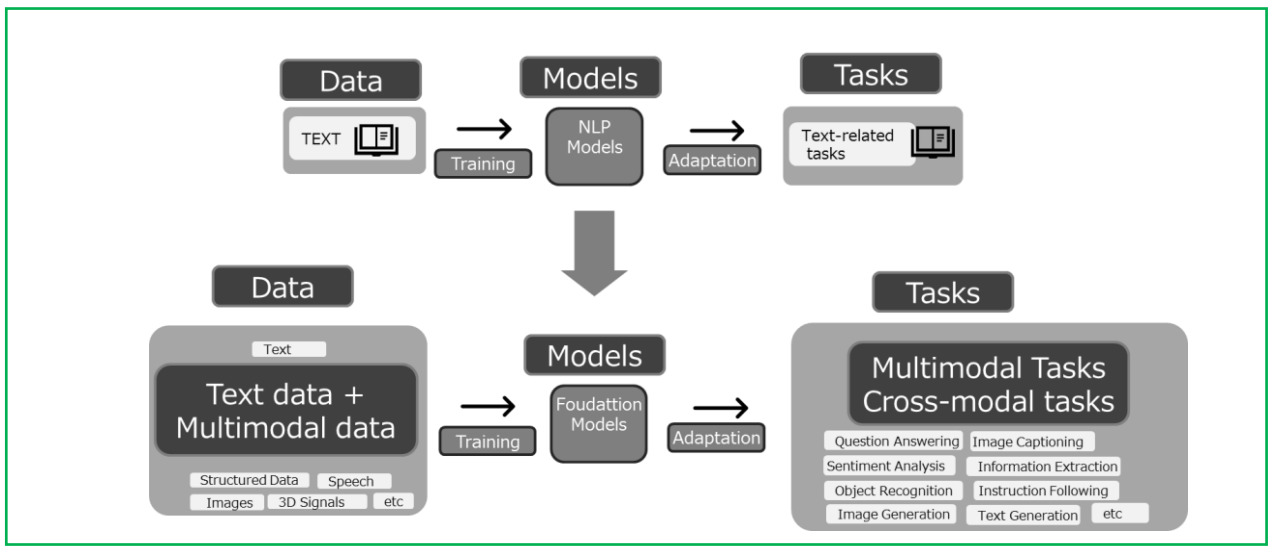

Fig. 8 From NLP models to Foundation Models[54]

Foundation Models significantly improve the NLP system further, which clearly functions in the way similar to humans. Moreover, the language systems they acquire or the learning process they use will likely to be completely different from those of human language learning. Understanding the meaning of the difference between NLP, which is obtained as the result of machine learning, and human language learning is extremely important for comprehending the limit and potentials of more powerful foundation models that will appear in the future, and for effectively utilizing them. For instance, human language acquisition appears to be extremely efficient compared to machine learning. For example, foundation models such as GPT-3 must learn language data that contains three to four digits more words than most people would hear or read in their lifetimes. No children would grow up in such an environment.

**4.3 Robotics and large-scale natural language model**

When building a large-scale language model, the model is trained with an extremely large amount of text data. The emergence of four arithmetic operations as a result of this was discussed earlier. Moreover, it is natural to consider the possibility of functions other than language emerging from the training data. In the following, discussions on robotics, which do not appear to be close to NLP at first, will be presented.

Fig.9 shows the representative input data of foundation models and the case where the same is applied to robotics. It becomes clear that robotics must function as though it was an integrated system.

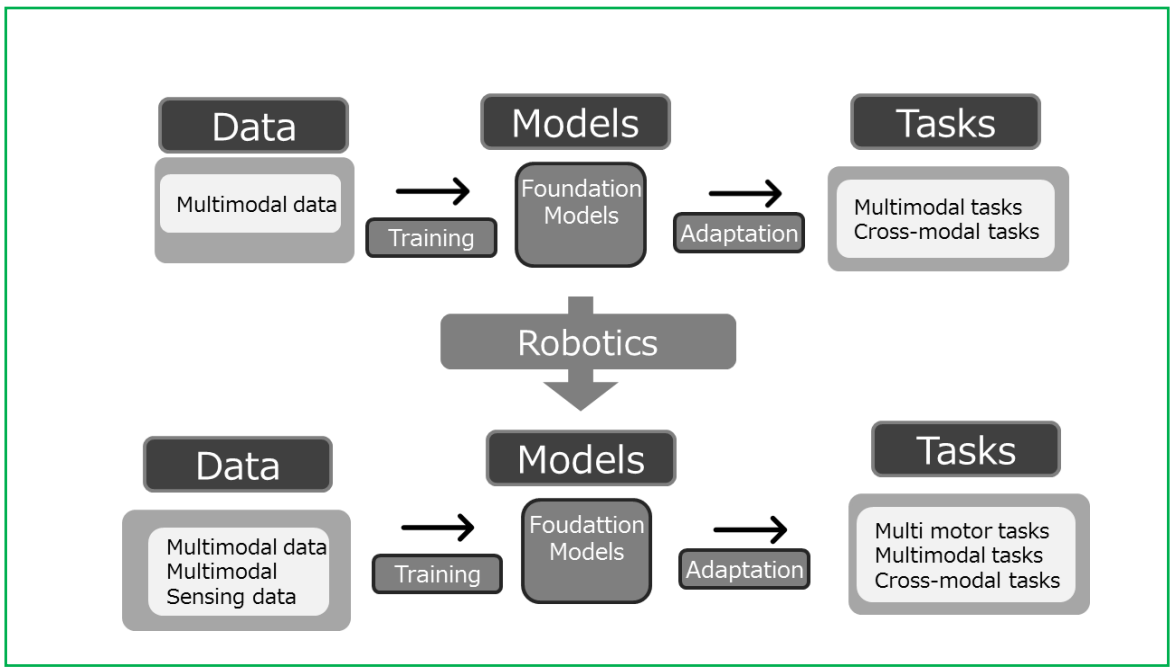

Fig. 9 Robotics as an example of using Foundation Models[54]

It is expected that whether foundation models can become literally foundational or not will be made clearer by considering their application to robotics. Furthermore, the fact that robots must interact with real environments is also very important. That is because it must clearly solve the symbol grounding problem, which is also often a problem for general-purpose AI.

Robotics requires large-scale datasets that cover diverse environments and operations. Currently, various approaches are being taken, including in regard to how data should be collected in real and virtual environments. Even without the application to robotics, simulations, interactions between robots, videos of humans, and explanations of operations/environments through natural language are all useful as data for foundation models. The fact that evaluations of success/failure from more complex data sources are required, and that the tasks involve the same operation under different environments, normally in zero-shot, significantly increases the difficulty.

Due to this fact, data with multiple modalities transcends the framework of NLP since they contain video. On the other hand, it is inferred that by taking into account the characteristics of the foundation model, a wide array of issues could be resolved. This includes specifications for robotics, supervised/unsupervised learning and reinforcement learning[65,66].

**4.4 Correspondence with the real world.**

Issuing orders, or inputs, to robots in normal language is one of the dream technologies. To achieve this, robots must understand the common sense and premises that supplement human conversation. Thus, considering input to robots as an extension of the aforementioned DALL・E and CLIP would be extremely useful. If the content of a text can be converted into images, it can potentially be turned into robot operations in a similar way. There is a report stating that the image capture generating capacity of CLIP is superior to models using deep imaging or room layout mapping in the Room Rearrangement Challenge[67]. This indicates that the arrangement of objects is accurately grasped. With this in mind, let us think of a general and ideal interaction with a robot.

Let us think of a situation where, while following an input stating "prepare breakfast," the robot infers/supplements all other elements to prepare said breakfast. A human would ask about unclear information on the spot to supplement it. Furthermore, he/she would first check what ingredients are available before preparing the breakfast. In addition, if one does not understand the concept of the place called the kitchen, one cannot prepare breakfast to begin with. To take real action, a robot must process numerous conditions and adapt to its environment. To solve this issue, it is necessary to integrate all the information and link the same to physical realization in response to text (or speech) input. This is significantly different in its character from the research problems of NLP or image processing (computer vision). At Stanford University, this possibility is being verified by using foundation models as the foundational models for robotics. This is one of the approaches to realize the usage of robots in everyday life.

In the high-level context of breakfast preparation, there is a significant amount of ambiguity, and it involves interdependent and interacting tasks. Furthermore, as it involves interaction with the physical world, every task requires details to be concretely defined. In order for a task to be sufficiently concrete and clear and to achieve its objectives, it must be broken into sub-tasks, and sub-objectives must be designated. Following this, the task to be executed can be specified. While natural language models are currently not able to realize every task, one can say that they have more than sufficient potential to be foundation models. Needless say, foundation models that can adapt to a body that interacts with its environment, like a robot, are more evolved general-purpose AI.

By understanding the purpose of a task, the robot will become capable of diverse conditioning accompanying a hierarchical task structure. As task specifications are normally provided by humans, quantifiable metrics for objectively measuring the task itself, to understand how a robot actually evaluates and completes the task, are not always available. Foundation Models function to convert the task specifications written in a natural way by humans into the task specifications for robots. This conversion includes a self-evaluation function through a reward function and signals necessary for operation. Thus, it will also become possible for the robot to evaluate the task by itself. They believe that, as a result, it will become possible to provide optimization of robot operation, diagnosis of obstacles and feedback to humans.

For instance, it is already known that it is possible to encode the general explanation of a kitchen, such as the facts that the oven is normally placed by a wall and it must be switched on to produce heat, into the prior information learned through the offline video of human interactions with robots, texts and/or simulations. Through such common sense knowledge, physical prior probability and visual prior probability, the robot is expected to become capable of adapting to new environments more efficiently. Similarly, by using many cooking videos in the training data set, it will become possible to learn how to adapt to the taste of a small number of users, or that of a specific one, from the foundational model of robot task training through adjustment of the data, in addition to learning the policies of common skills such as "frying an egg." With a trained foundation model, one can simply provide a description saying "I want to eat breakfast," and then it will prepare a fried egg cooked in the way one likes, as well as bread and coffee.

From another perspective, in the application of foundation models to robotics, there are major issues accompanying physical realization, namely safety and robustness. As mentioned earlier, this is the perspective that makes it significantly different from NLP or computer vision. Needless to say, what is ultimately necessary for a robot is its operation in a real environment, regardless of how satisfactory its performance in a virtual space is after training with multimodal data. On the one hand, it is extremely important to ensure safety and robustness during this operation. On the other hand, this would make development of a foundational model even more complicated. That is because this perspective is not necessary contained in the data. Another problem is the fact that the data collected in real environments contains information that does not exist in foundation models, because its collection involves direct interaction with environments in physical worlds. It is not difficult to imagine that such information will be closely related to safety. This is normally handled by restricting the system to limit its interaction with the real world. The task is learned under this environment without making critical mistakes.

There is often debate about the chicken or the egg problem, where safety limitations for the system must be assigned before data can be collected. It is imperative to consider many factors when interacting and training in an environment. Take training in the kitchen as an example. In the kitchen, the system must learn to detect anything that can break. One also has to consider whether items that may break should be replaced when collecting data. Alternatively, the environment should be left as it is so that the system will learn that these items can be damaged. When conducting training without interruption, one must choose the former. However, the latter is necessary to generalize various stimulations and unpredictable behaviors. The causal analysis of agents[68], safety evaluation tools and realistic simulation environments[69,70,71] are being debated.

Thus, one can conclude that robotics and foundation models have an extremely close relationship. At the same time, data is even more important in this field. Significant effort is required to collect or generate data that covers the wide range that includes diverse environments of the physical world and various states of execution. Moreover, this data is directly connected to the safety and robustness of the system. This offers a very important perspective.

However, it is not as simple as the adaptation of foundation models by robotics leading to the realization of an ideal robot. For instance, in order for a robot to reproduce the action of closing buttons while putting on clothes, a foundation model alone is insufficient. That is because, while the foundation model would contain the expression "closing buttons," the specific procedure would not be outlined. To be exact, the action of closing one button would involve holding a button, holding the edge of the front side of one's top (crest) corresponding to the position of the button, inserting the button held at right angles into a buttonhole from the back, and then making the button horizontal again when releasing it. No such descriptions would be contained in normal text data. Folding clothes is a similar case. There is a possibility that foundation models are not actually suitable for actions humans perform without thinking. One can only hope that this will be addressed by future developments.

## 5 Introduction of the topics related to natural language models after GPT-3

In this section, the technologies released after GPT-3 will be briefly discussed. These technologies are directly or indirectly linked to GPT-3.

### 5.1 MT-NLG[72], a powerful natural language model by Microsoft and NVIDIA

MT-NLG is a natural language model developed by Microsoft and NVIDIA, and it has 530 billion parameters, which is 3 times more than those in GPT-3. Consisting of 105 layers, it is currently the largest natural language model. At release, it was announced that its capability exceeded the performance of all existing NLP models. Training with such a large number of parameters required coordination with hardware, and realization of such was only possible through collaboration between Microsoft and NVIDIA. Its heart is a super computer consisting of 560 DCG servers, each one equipped with eight A100 GPUs that contains 432 tensor cores for machine learning and 80 GB RAM.

The size of its training dataset is 1.5 TB, comprising several hundred billion text data obtained from 11 databases, including Wikipedia and PubMed.

It has been reported that MT-NLG scored high on the benchmarks for prediction of complete text according to its meaning, reading comprehension, generation of logical conclusions, generation of conclusions in natural language and distinguishing between words with multiple meanings.

Moreover, the developers commented that they identified an emergence of the function that displayed the comprehension of the simplest mathematics, which was also reported in other large-scale natural language models[73]. Nevertheless, the bias problem exists in MT-NLG, and scaling up alone will not solve it. Below is an example from MT-NLG. It is interesting that "True" or "False" judgments are being made, indicating that it is capable of evaluating the content of the text.

Prompt:
The banker forgot that the judge advised the professors.
Question: The judge advised the professors. True or False?
Answer:(MT-NLG)True. The banker forgot that the judge advised the professors.

Prompt:
Famous professors supported the secretary.
Question: Professors supported the secretary. True or False?
Answer:(MT-NLG)True. The secretary was supported by famous professors.

### 5.2 Microsoft Project Florence-VL[74]

One of the chief goals of artificial intelligence is to develop an algorithm that can effectively learn from multimodality data. The linking of image and text is equivalent to the linking of vision and language we habitually perform to understand the world around us. At the moment, bidirectional conversing between text and image, namely searching images that correspond the best to a text query and generating a text explanation of an image, is becoming common. One approach to this is Florence-VL (Florence-Vision and Language Project) launched by Microsoft.

Web and social media are full of image and text pairs, which functions as an annotation of the image. The corresponding can be used as data labels for supervised learning. Moreover, as many videos contain audio channels explaining their contents, it is possible to turn such audio into text, which can then beused as labels. If such data can be used for Vision-Language Pretraining, it will be possible to distill cross-modal knowledge[75] in the form of texts, images, and audio files, where knowledge learned in one modality becomes useful for learning in another modality, and manual data labeling is no longeer required.

For Florence-VL, a large-scale model was prepared through pretraining consisting of the prediction of masked elements based on their context. As was the case with existing Transformer-type models, self-supervised learning using a large number of image-text pair data from the Web and social media was employed. The cross-modal representation of this model was fine-tuned to accommodate so-called downstream tasks. Microsoft presented the result of this project in a series of papers that were published one after the other. UNITER, OSCAR, VILLA, VinVL, VIVO, and TAP all have their own unique features. For instance, VIVO achieved equivalency with humans in the new image caption task (nocaps). After strengthening its pretraining by using the scene text detected in images, it achieved No.1 in the TAP TextCaps Challenge2021.

Moreover, UFO and METER were developed for the end-to-end pretraining necessary for further improvement of Florence-VL. In addition, UNICORN, an the integrated visual image model, LEMON, and SimVIM related to scaling and PICa for multimodal few-shot learning have been developed. How these projects are closely linked to each other with the ultimate goal in mind can be demonstrated as thus. For instance, the model architecture developed through end-to-end pretraining functions as a component of the integrated visual language modeling can be increased and turned into an integrated solution by using another method developed for upscaling. This would naturally enable usage of the few-shot learning function. As a result, and by using several examples in the context, an integrated VL foundational model is born, and it can easily adapt to various tasks.

## 5.3 BigScience T0 [76]

As mentioned many times in this paper, it has been reported that recent large-scale natural language models display generalization through zero-shot learning in diverse tasks[77]. A hypothesis proposing that this is the result of multitask learning in natural language model training has been put forward[78]. How does generalization in zero-shot learning work? How does multitask learning induce it? To conduct a large-scale test for answering these questions, T0, a system that can perform simple mapping of common natural language tasks in a prompt format humans can read, was developed and used for detailed analysis, and the result was published. It converted large-scale supervised datasets, and several prompts using various natural languages were prepared for each dataset.

The prelearnt encoder/decoder model[79,80] is fine-tuned using this multitask prompting that covers diverse tasks[81].

The result successfully answered unknown text questions. This means zero-shot learning was achieved. The zero-shot performances after the training using several standard datasets displayed many results that were superior to other models, whose sizes were up to 16 times larger than this model. Moreover, it achieved performance superior to other models, whose sizes were up to six times larger, in the subset of BIG-Bench benchmark tasks. It demonstrated the potential for decreasing the model size through prompt learning.

**5.4 LaMDA, dialog approach of Google[82]**

The LaMDA (Language Model for Dialog Applications), which Google developed with a focus on conversation, saw further evolution in the field of conversation. Transformer, which was invented by Google, learns the connections between words. Thus, it predicts the subsequent words in a sentence with high precision. However in conversation, that is insufficient, and it often has to understand the hidden nuance in each word before performing its prediction. Therefore, LaMDA employs a method to learn from conversations.

To appropriately respond during a conversation, it is insufficient to simply return a response that corresponds to the words being used. Instead, it is necessary to understand the context of the topics being discussed in the conversation and then offer a concrete response that is specific to this context. As Google itself explained, the probability of a response saying "that is good" being appropriate is considerably high. However, as is the case in human conversation, it is insufficient to continue a conversation through such unspecific responses, and evaluation of how vibrantly the ongoing conversation is going is also important for that conversation to develop and expand.

A conversation does not have a landing point, i.e., a final destination, at first, and its topic may change significantly toward the end. Such a situation is sometimes praised as being full of insights, or witty. Regarding this, LaMDA learns the relevancy between topics, which cannot be learned only through word-level relationships, through conversation learning. Naturally, it is starting to examine how to offer convincing answers that also include fact checking, which has become a problem in recent years. It states that it will continue the effort to integrate conversational ability into more products.

LaMDA2 was introduced during the developer summit in May 2022. LaMDA2 was refined through trials by a few thousand staff members of the company. One of the function demonstrations of LaMDA2, named 'Imagine It,' is capable of expansion into clearly relevant topics even without previously defined answers directly related to human input. For instance, it can expand the conversation in this way: Deep sea → (Image of) Mariana Trench → Creatures, submarine… through questions that stimulate the dialog. This is the result of the model integrating and responding to the training data. They declare that backtracking never occurs in the conversation, and it is capable of continuously expanding dialog.

## 5.5 DeepMind RETRO (Retrieval Enhanced TRansfOrmers)

With conventional large-scale natural language models, the size of the model and data is related to the performance of each, meaning that a larger size will lead to better performance. In response, RETRO successfully maintained the performance of the model at a smaller size through direct retrieval of training datasets by utilizing search. As a result, it realized significant performance improvement compared to the standard Transformer-based model with the same number of parameters.

Training of a natural language model requires vast computational resources. For the training of large-scale natural language models, such as GPT-3 with 175 billion parameters and Microsoft with 530 billion parameters, the computing power required is too large. Thus, RETRO uses an external database to reduce the training cost without shrinking the training sample size. In other words, it outsources some of the parameters, which the model is expected to have within it, to an external database. This can be understood through simplification of the Transformer mechanism. The weight of the context of the word in focus is normally heavy. Therefore, it is obvious that the best example of a text with heavy weight in relation to the word in focus is the text that contains that very word. Thus, it is logical that the model functions sufficiently when it uses the text data itself for the weight of words neighboring the word focus while using the training in Transformer for farther relationships. This reduces the number of necessary parameters, and sufficient performance can be achieved through training with fewer computing resources. The above explanation uses an extreme example. In reality, they devise adjustments because the neighbors cannot be ignored completely.

RETRO trains its model using a dataset composed of news articles in 10 languages, including English, Spanish, German, French, Russian, Chinese, Swahili, and Urdu, Wikipedia texts, books, and GitHub texts. The RETRO neural network only has 7 billion parameters. Therefore, it employs a method whereby they are supplemented by a database that contains approximately two trillion text passages.

When RETRO generates text, it performs a nearest neighbor search for each segment (equivalent to a paragraph of a text), and returns a similar text found from its training database and results that continue the aforementioned text. These arrays are useful for predicting subsequent texts from the input text. The architecture of RETRO combines the common Self-Attention at the text level and Cross-Attention with neighbor search at the finer passage level. As a result, it can realize more precise continuation based on facts. Furthermore, RETRO improve results through the direct control of search databases to increase the interpretation possibilities of model prediction and improve the accuracy of texts. It uses the database to search for passages that are the same as the written text and compares them. This enables it to perform more precise prediction. The performance of the natural language model continuously improved as the size of the search database increased. At least up to two trillion tokens have been verified, and the two trillion tokens of training data was used as an external database as is.

In the experiment using Pile, which is a standard natural language model benchmark, the RETRO model with 7.5 billion parameters displayed performance exceeding that of Jurassic-1, which has 175 billion parameters, in 10/16 datasets, and displayed performance exceeding that of Gopher, which has 280 billion parameters, in 9/16 datasets. Moreover, satisfactory performance can be achieved at 1/16 the parameter size.

Moreover, the database is capable of updating the neural network without relearning, thereby demonstrating that it is capable of the speedy addition of new information and deletion of old or incorrect information. DeepMind claims that external memory systems such as RETRO are more transparent than black box models such as GPT-3. WebGPT[83] by OpenAI employs a method to reuse the Web as the database.

**5.6 South Korean LG EXAONE[84]**

LG released EXAONE, a giant AI comprising a large-scale natural language model produced in collaboration with Google, is also the largest AI in South Korea. As LG is a conglomerate engaging in diverse businesses, employing various AI applications centered on a natural language model, which is the concept outlined by foundation models, makes sense as its corporate strategy. One can say that EXAONE plays a central role in this strategy.

It uses a text corpus comprising 600 billion phrases and more than 250 million images as its training data, which combines these two types of data. It uses 300 billion parameters. In the explanation by LG, EXAONE is claimed to be an AI that learns big data by itself, and is capable of thinking, learning, and making decisions like humans, which can be applied to diverse fields without being limited to specific purposes. Needless to say, EXA is also the prefix "EXA" meaning 10 to the power of 18 (i.e., a quintillion). If all the words humanity ever used were saved as data, its size is estimated to be approximately 5 Exabytes. LG explains that the name EXAONE implies this idea.

LG AI Research changes the parameter number of the neural network in stages. It has been increasing its size in stages from 1.3 billion to 13 billion, 39 billion, and 175 billion to study its characteristics. Its result and previous reports suggest that, in theory, training of AI based on deep learning would be more refined when there are more parameters. In multimodality learning, which is mutual learning of texts and images, a larger number of parameters is clearly an advantage. EXAONE achieved this through large-scale parameters.

Similar to other models, EXAONE creates a hat with a pumpkin pattern if you tell it to "make a hat with a pumpkin pattern." LG AI Research declares that EXAONE reaches the state where such output is made through inferences and creations that exceed the level of mere understanding of input and data. LG AI Research published a video of EXAONE preparing a Christmas party inside a metaverse space. The video shows the process whereby it understands the intentions outlined by the client, makes judgments and creates the decoration.

According to the plan of LG AI Research, EXAONE is expected to become the "specialist AI for the top 1% standard" in virtually every field, including manufacturing, research, education, and finance. First, it opened API to the companies within the LG group to encourage them to use EXAONE. This enabled companies with the LG group to use the giant AI for every business field of LG, including electronics, chemistry, and communication. It appears that each company is actually using EXAONE and achieving results, including the advancement in the performance of chatbots and the extraction of new materials/new substances through analysis and learning of approximately 20 million publications in the chemistry field from last 100 years.

LG group is in close collaboration with Google toward the completion of EXAONE. The cutting-edge AI chip "TPUv4," which Google released on May 2021 and is not yet commercially available, is used to develop EXAONE. Moreover, LG is collaborating with Google Brain for its software. LG has stated that the future goals for EXAONE include extension of its parameters to 600 billion, as well as the formation of global allied forces of national and international AI to create a colossal AI ecosystem through collective intelligence; indicatingLG belief that AI strategy will play the central role in its future development. Google Brain is collaborating with them to achieve its goal of breaking nvidia's stronghold. In fact, cooperation with LG is significant for Google. In AI learning, NVIDIA has more than 80% of the market share, representing a *de facto* monopoly. If LG manages to create a B2B business model built upon its affiliated companies in the infrastructure field, Google would obtain a powerful reference for challenging NVIDIA.

**5.7 Chinese BAAI Wu Dao 2.0[85]**

The Beijing Academy of Artificial Intelligence (BAAI, Chinese name: 北京智源人工智能研究院) released a pretrained deep learning model Wu Dao2.0 with an astonishing 1.75 trillion parameters, which is described as "the first in China" and "the biggest in the world." They stated that, in addition to the simulation of conversational speech, writing of poetry and comprehension of images, it can even generate recipes, and everything NLP models could do in English is now possible in Chinese. The model was trained with 4.9 TB of images and texts, and this includes 1.2 terabyte of texts in two languages, namely Chinese and English. WuDao 2.0 already has 22 partner companies, including the smartphone maker Xiaomi and the major short video company Kuaishou.

BAAI has reported that this natural language model has achieved the state-of-the-art (SOTA) level in nine benchmarks, and even exceeded it in some of them. When they released Wu Dao2.0, they also unveiled Hua Zhibing, the first Chinese virtual student in the world. Hua is said to be capable of learning by herself, drawing pictures and writing poems. During the presentation, it was declared that she will be able to learn how to code in the future. These are unique abilities not found in the original version of GPT-3. However, it has been reported that GPT-3 also becomes capable of generating codes after few-shot learning. While the details of the training data of Wu Dao are unknown, it is possible that the difference in the content of the training data is reflected in its early functions.

### 5.8 OpenAI GLIDE, DALL・E2

CLIP is capable of reading the relative positions of objects in an image because it assigns detailed captions to images. One of the problems it has is that there is a limit to its capacity when dealing with more abstract expressions or certain types of tasks, such as counting the number of objects in an image or measuring which car is the closest one from the camera in an image with several cars. The answers it gave to these tasks were far from accurate, only slightly better than random estimation. Another problem was that it could not give satisfactory results, even with more detailed classification (such as car model categories and flower classification).

While CLIP is an unprecedentedly powerful zero-shot classifier, there are occasions where it cannot cope with changes in word choice or turns of phrase. When this occurs, trial and error through prompt learning becomes necessary, requiring it to respond to each case individually. This necessitated its reexamination, as well as that of GAN[86], which is its conventional generative model.

GAN is a generative model with extremely good ability, and numerous examinations were performed to realize satisfactory generative function. Currently, it requires advanced tuning of its hyperparameters and regularizer. This means that it is extremely difficult to apply it to domains other than the one it already learned. Thus, there is always a trade-off between significant quality improvement within its scope of application and considerable quality decrease outside this scope. The diffusion model was proposed as a solution for this problem other than GAN.

OpenAI has reported that the improved version of this diffusion model achieved better performance than GAN[87]. While GAN uses noise vectors when generating images, in principle, there is no guarantee for these generated images. This is the reason why it requires the aforementioned tuning. To improve this situation, the diffusion model gradually adds Gaussian noise the original image data, replacing the data with noise until the original data is completely lost, replaced entirely by noise. This procedure follows the Markov process. In the training of the generative model, the data is restored in reverse while removing the noise. This means the control of generated images, which was difficult for GAN, could be learned from the beginning to the end through the relationship between noise and image.

GLIDE applied this diffusion model to the model intended for text → image. To be specific, it employed a diffusion model with 3.5 billion parameters using a text encoder, whose condition is natural language description, for its training. GLIDE received an 87% more positive evaluation in being photo realistic, and 69% more positive evaluation in caption similarity, than DALL・E, its preceding model. Rendering of these images can be performed in zero-shot. However, the cases where generation of photorealistic images was difficult when the texts used were complex were identified. In response, the model was fine-tuned to add an editing function, making it capable of painting parts of an image, to enable it to generate high-quality images from more complex text. When a part of an image is painted over and then text is entered, the part that was painted over changes according to the text. Moreover, this changed part contains shadows and reflections that match the style and lighting of the surrounding context, enabling synthesis that can blend in with the surrounding.

DALL・E 2[88] was released on the 6th of April, 2022. Thanks to unCLIP, which adapted the diffusion model, its significant improvement from DALL・E is immediately apparent. An example of an image generated by DALL・E 2 through text input immediately shows that the text content and the image data representing it are more detailed. The evolution of its image generation function from text is so advanced that one cannot help wondering if the model acquired imagination and creativity. Furthermore, unCLIP is expected to have a representational ability that is superior to that of CLIP. For instance, it is highly likely that it will be able to generate a perfect digital twin and world model only from a few images. Its future examination will be drawing attention.

### 5.9 MLP[89] Reexamination, beyond Transformer

Google has proposed gMLP[90], a network architecture only based on MLP with gating. It is the improved version of Transformer without the Attention mechanism. It demonstrated the same level of capability as that of Transformer in main languages and in the field of image processing. In recent years, detailed analysis of why MLP such as gMLP can achieve high capability has started. For instance, there is a heated discussion on whether it was due to the effect of Attention. For instance, if Attention can be regarded as the introduction of dynamic parameter (effective) bias, MLP can be regarded as static parameter bias.

How the way bias (weight of the attention point) changes performance is currently a very important subject. MLP, which is the focus here, realizes static bias through SGU[91]. While Attention clarifies the reasoning for the parameter weight by following the context, in another word connections, computation of SGU is based on simple correlation. This is the significant difference, though admittedly, this is not the most precise description. In images, their difference is especially conspicuous, and naturally, computation is easier in the latter. The high computation cost is the issue with Transformer that is often mentioned. Thus, this is attracting attention as a possible solution for improving the performance while limiting the computation cost. This is one of the reasons why the performance can be scaled through the size of data and number of parameters. Regardless of whether it is explicit or not, how to implement it while considering the weight relationship between the target vector and other data vectors will be extremely important.

Other perspectives include the comparison between gMLP and ViT (Vision Transformer)[92], which is the application of Transformer to images. One of the focuses of its comparison to other methods is how the basis for image recognition is acquired. The same format is derived for the basis for the features of human vision regardless of whether biological analysis[93], DL or frame rate of tensor analysis[94] is used. Therefore, it is likely that no problem would be caused by thinking that the general basis for seeing things and distinguishing between them is limited. For instance, there is a report stating that the letters and alphabets we use belong to an even more simple basis group[95]. Thus, it may not be necessary to acquire them through training with large amounts of data.

The network structure of CNN in the imaging field is designed so as to explicitly derive the basis. How is this basis, or the feature filter that conforms to it, extracted and structured? In Transformer, what came from where is traced by the position embedding layer, while SGU does not perform this process, and uses correlation instead. This difference is also reflected in the computation cost.

Interestingly, in addition to the basis observed in the aforementioned visual cortex of the brain (low-level visual cortex), correlated bases composed of several figures were also observed in the visual cortex of the brain (high-level visual cortex)[96]. As the brain is a super efficient computation system running at 20W, it is inferred to make effective use of various features. Thus, it is believed to be adaptively utilizing various methods including the future development, instead of only using one of the methods that are currently deemed to be effective such as CNN, Transformer, or gMLP.

Recently, there have been reports on the generation of CNN by HyperTransformer[97] and on increasing the speed of MLP through hash calculation. We are currently in a situation where Transformer, which was initially used for NLP, is showing new possibilities in the CV field[98], and is expected to spread into other technological fields.

MetaFormer[99] is proposed in the report, which suggests that good results can be obtained by using only Transformer without assigning a token mixer module, typified by Attention. MetaFormer is an abstraction of Transformer whose token mixer module has not been defined, whereby the token mixer becomes a kind of model architecture that can be exchanged, for instance, with spatial MLP.

For instance, when it is exchanged with a simple spatial pooling operator, it becomes PoolFormer. PoolFormer uses a simple pooling operator as the mixer. Naturally, the pooling operator only functions to aggregate a token with tokens nearby and average them, and it does not mix information. Moreover, in PoolFormer, the pooling operator does not have any parameters it can learn. The fact that it nevertheless displayed high performance in computer vision tasks such as ImageNet-1k and DieT-B / ResMLP-B24 is surprising.

What has been discussed thus far is the analysis of Transformer that was derived from the application of Transformer used for NLP to image processing. In the same vein, there are reports that question whether Attention is essential, even though it is effective for some tasks in NLP.

Furthermore, there is a recent discussion on intermodal training methods. In addition, data2vec[100], which vectorizes all data, images, audio, and natural language in the same way, has been released. It acquires latent representation, regardless of the data type, through the two modes of Teacher mode, which conducts feature learning through complete data, and student mode, which predicts complete data from masked data. It analyses which layer of Transformer to be used for masked prediction in detail. The result was SOTA or its equivalent at each conventional modal. What comes after Transformer is already being explored, and 2021 can be deemed as the first year of this exploration.

## 6 Important Challenges

Earlier it was mentioned how the biases contained in natural language models often become a problem. Let us discuss this point further. Equality and discrimination often surface as social issues. It is known that a natural language model is heavily influenced by its training data, and how it can be corrected so as to avoid these problems is frequently debated. As a precondition, an algorithm must be balanced. However, one must be aware that bias is still possible. For instance, if LGBTK+ ID terminologies are excluded from the training data, or if their number is very small, they could be underestimated, or completely deleted, from the data.

If an algorithm is to be neutral, this is in fact a correct operation. However, this would be a major issue for a model. The relationship between training data and the inherent bias in a natural language model is not accurately known. Therefore, it is currently believed that if a scaling law that can be applied to biases is established through small-scale and systematic research, it will also be possible to apply it to large-scale data practices. The fact that every natural language model must use neutral datasets that nevertheless guarantee equality is a future challenge.

Google conducts especially strict studies on whether it is adhering to its own AI principles. In their services, language plays a central role, and many mechanisms for avoiding the misuse of language have been introduced, based on the premise that language is one of the best tools humanity has. They are required to be especially vigilant toward prejudice caused through misuse, malicious expressions and the reproduction of misleading information. For instance, with a conversation application, careful examination of its results (i.e. conversation outputs), does not exclude possibilities for misuse of the model itself. Such misuse must also be prevented.

For this reason, Google states that the priority when creating technologies, such as LaMDA, is to sufficiently discuss the actions that can minimize the risks and then to execute the same. The research and development of these technologies over many years has addressed a number of related problems, such as how inequality bias affects machine learning models. Furthermore, based on these experiences, they provide an open source resource for analyzing models and training data, and they accept open, external examinations. For instance, regarding LaMDA, they explain that it was carefully examined at each stage of its development. They are attempting to demonstrate their attitude toward equality in their natural language models and natural language applications.

Efforts to address content that is incorrect, inappropriate, and/or offensive were also emphasized during the Google developer summit in 2022, where they presented how they are actively utilizing feedback from internal users. Google also reported that they are accepting methods that follow repetitive and regular principles as well as the opinions of external experts, in accordance with their AI ethics.

In May 2022, Google published a report on Imagen[101]. It is the Google equivalent of DALL・E2 that generates images from texts. It employs the diffusion model and has updated SOTA in several categories. Its comparison to DALL・E2 undoubtedly shows more refined image generation results. However, Google announced that they will not publish its demo or code due to the following reasons. It employed widely used datasets for the image data in its training data (e.g., LAION-400M[102]). This set is known to contain many toxic images. While they, of course, carefully removed toxic data, the curation of Web level text data used for its language model was still insufficient. Therefore, they reflected on how the risk of harmful stereotypes and expressions being encoded still remains. On this occasion, Google intentionally included toxic expressions in their evaluation. This demonstrates the honesty of this report in both the state of research and development and the bias problem.

OpenAI also released InstructGPT[103], which improved the text generation by GPT-3 to better understand the intentions of its users, as well as to process toxic biases. For InstructGPT, the method called RLHF[104], where humans are involved in reinforcement learning, is employed. By using human preference as a reward signal, it is possible to fine-tune the model. As the problems in text generation are complex and subjective, it plays a significant role in improving situations with which the simple and automatic metrics that exist today cannot fully cope.

## 7 Conclusion

This report summarized recent debates, with a particular focus on developments originating from NLP models. It demonstrated that today, it is possible to say that large-scale natural language models are already at the center of general-purpose AI. It is one of the representative AI fields, and we must continue to pay attention to it. At the moment, super computer level computational resources, and larger and larger amounts of data, are required for large-scale models. Thus, powerful players are becoming more powerful, and it is almost impossible for new players to enter this field. At the same time, it is also true that academic exploration of upcoming computation models in small-scale environments will become more and more important. The fact that Attention (Transformer) was only a result of small-scale research and development at the time of its initial release proves this point. To be implemented in society, developments must be conducted while considering how to use existing models, which, fortunately, are most often already available.

In a recent DeepMind report[105], it was shown that the relationship between the parameter number of a model and the scale of its training data is not necessarily linear. In fact, it was shown that a model with 70 billion parameters exceeded the performance of a model with 175 parameters in almost every benchmark. Moreover, it even exceeded human scores in some benchmarks. Furthermore, Flamingo[106], with 80 billion parameters, was announced as a visual language model in late April 2022. Its image contents description capability is even more refined.

Google published a report on the Pathways Language Model (PaLM)[107] with 540 billion parameters on April 4th, 2022. According to the report, it updated SOTA in 28 out of 29 NLP benchmarks. What was particularly astonishing was the fact that it could understand jokes and explain the points of those jokes, something with which previous models struggled. It adapted a solution through multistep prompting called Chain-of-Thought Prompting to cope with a context that requires several stages of understanding. By supplying it with several question-and-answer pairs, it can learn a series of hierarchical structures. As a result, it became capable of hierarchical and staged understanding, enabling it to, for example, explain jokes, provide answers for a mathematical problem in several steps and generate more complex program codes.

Moreover, when seen from the perspective of the law of scalability[16], the achievement of PaLM suggests that efficient learning remains a possibility when the number of parameters is increased. We are only starting to understand the number of parameters, scale of training data and performance of models. Thus, it is a field where future developments and deepening discussions are expected.

In late April of 2022, Adept[108] was established. It professes Useful General Intelligence to induce further evolution of AI by building upon the developments in this field thus far. Its members are well-known for NLP and Transformer, typifying this fast-moving field. It appears that one would miss major opportunities unless one is able to at least follow these movements. Recent engagements with general-purpose intelligence include GATO, released by DeepMind on May 12th, 2022. It is declared to be a Generalist Agent[109]. Like other large-scale language models, it was trained with Transformer. It is an agent that functions with a single model, and performs the role of a multimodal, multitasking, and multibody generalist. In response to various tasks, including Atari, image captioning, chat, and playing stack block with a real robot arm, a single model can supply output suitable for each context, including text, movement of joints, and pushing buttons. The term generalist is indeed appropriate. This is a landmark achievement of research and development, and it gives the impression that we are one step closer to the realization of foundation models.

Lastly, let me add that interesting recent engagements show that the application of Transformer is starting to cause major innovations in fields unrelated to language. DeepMind has triggered major innovations in the structural analysis of proteins through its AlphFold[110]. In addition, AlphaFold2[111] brought even further development. While AlphaFold is based on CNN, AlphaFold2 is based on Transformer. This was not a coincidence, and it is natural to conclude that it was due to the high potential of the Transformer series. Needless to say, it is also true that as the task to convert the names of chemical substances into three-dimensional chemical structures involved symbols and structures composed of them, it had compatibility with Transformer.

Similarly, there are high expectations for Materials Informatics (MI), which is a data-driven technology to optimize/accelerate material development[112]. Regarding training using data, this field has its own unique problems, namely how to effectively use the data that is much smaller than text data, and how to precisely predict the structures and properties of new materials from extremely small data. In recent years, it has becom clear that pretrained models are also extremely effective for MI. Thus, it is predicted that pretrained models will continue to spread to various other fields.

<div style="text-align: right">Masahiro Yamamoto, Adviser of General Affairs Planning Department</div>
<div style="text-align: right">[Inquiries]</div>
<div style="text-align: right">Information-Technology Promotion Agency</div>
<div style="text-align: right">General Affairs Planning Department</div>
<div style="text-align: right">Email:  ga-airesearch@ipa.go.jp</div>

---

[1] Natural language processing was discussed at the Dartmouth Workshop in 1956, where the first discussions on artificial intelligence took place.
[2] https://www.ipa.go.jp/digital/chousa/trend/ai.html
[3] GPT-3: An abbreviation of Generative Pre-Training-3, which is the third generation of the general-purpose language model developed by Open-AI. https://beta.openai.com/docs/engines/gpt-3, https://openai.com/blog/openai-api/
[4] Google Duplex: A.I. Assistant Calls Local Businesses To Make Appointments
https://www.youtube.com/watch?v=D5VN56jQMWM
[5] Duplex is getting smarter and making life a little easier
https://blog.google/technology/ai/duplex-helpful-updates/
[6] Bringing you the next-generation Google Assistant
https://blog.google/products/assistant/next-genenation-google-assistant-io/
[7] Language model: A machine learning model for natural language processing. Models based on probability are now mainstream
[8] BERT: Bidirectional Encoder Representations from Transformers. The name of the machine learning technology for language models that Google introduced to its search engine in 2019.
[9] Parameter(s): Variable(s) to be set in machine learning.
[10] Prepared independently from the following data source;
https://www.youtube.com/watch?v=G5lmya6eKtc, https://ja.stateofaiguides.com/20200914-future-of-nlp/
[11] H100 is also an NVIDIA GPU, and it uses the latest Hopper architecture released during GTC2022.
https://www.nvidia.com/ja-jp/gtc/keynote/
[12] Language modelling at scale: Gopher, ethical considerations, and retrieval,
https://www.deepmind.com/blog/language-modelling-at-scale-gopher-ethical-considerations-and-retrieval
[13] Language Models are Few-Shot Learners. Tom B. Brown et al. 2020, https://arxiv.org/abs/2005.14165
[14] GPU: Graphics Processing Units. When this is used for general-purpose calculation, it is called (General-purpose computing on graphics processing units.
[15] RNN: Recurrent Neural Network. It has a network structure where a part of the output is connected to input. Before Transformer, LSTM, which is a type of RNN, was the mainstream in natural language processing.
[16] Attention Is All You Need., A. Vaswani et al. 2017, https://arxiv.org/abs/1706.03762
[17] Machine learning can be broadly divided into supervised learning, unsupervised learning, and reinforcement learning, and training data is necessary for the first. For details, please refer 2021 DX White Paper Appendix, Section 1, AI Technology, 3. Learning. https://www.ipa.go.jp/files/000093703.pdf
[18]BERT: Pre-training of Deep Bidirectional Transformers for Language Understanding., J Devlin el al. 2018, https://arxiv.org/abs/1810.04805
[19] While Transformer was initially a machine learning method for natural language processing, its effectiveness when applied to images has also been demonstrated. This field is known as Vision Transformer (ViT).

[20] Scaling Laws for Neural Language Models., J. Kaplan et al. 2020, https://arxiv.org/abs/2001.08361
[21] This indicates learning with a large-scale corpus for the purpose of general-purpose language skills acquisition. It must be conducted in advance for the training for the intended task. This is expected to enable high performance, even when the data for the intended task is small.
[22] Language Models are Few-Shot Learners., T.B. Brown el al.2020, https://arxiv.org/abs/2005.14165
[23] The method by which the learner is taught using short exchanges of sentences in dialogue format.
[24] https://paperswithcode.com/area/natural-language-processing
[25] Dynabench: Rethinking Benchmarking in NLP., D. Kiela et al. 2021, https://aclanthology.org/2021.naacl-main.324.pdf
[26] MNIS: Dataset widely used as the benchmark for handwritten text recognition. https://www.vision-systems.com/home/article/16737424/support-vector-machines-speed-pattern-recognition
[27] GLUE: https://gluebenchmark.com/
[28] https://ruder.io/nlp-benchmarking/、https://github.com/kwchurch/Benchmarking_past_present_future
[29] There are following examples. Example where Microsoft from the United States apologized for a comment by its AI (https://jp.reuters.com/article/tay-idJPKCN0WU056), example of gender discrimination by the AI tool used by Amazon of the United States. (https://www.reuters.com/article/us-amazon-com-jobs-automation-insight/amazon-scraps-secret-ai-recruiting-tool-that-showed-bias-against-women-idUSKCN1MK08G)
[30] Google I/O keynote lecture, https://io.google/2022/program/8e80903f-955f-4a5b-9118-b0ce4acdb0e6/intl/ja/
[31] https://www.ethnologue.com/
[32] For example, https://maraoz.com/2020/07/18/openai-gpt3/
[33] https://gpt3demo.com/
[34] One can verify this by entering GPT-3 as the name of the author at https://www.amazon.co.jp/advanced-search/books
[35] https://www.othersideai.com/
[36] https://www.usebroca.com/
[37] https://snazzy.ai/
[38] https://www.forethought.ai/
[39] https://www.gong.io/
[40] https://www.moveworks.com/
[41] https://www.observe.ai/
[42] DALL・E: AI for generating images from texts developed by OpenAI. Zero-Shot Text-to-Image Generation., A. Ramesh et al. 2021, https://arxiv.org/abs/2102.12092
[43] https://openai.com/blog/dall-e/
[44] CLIP: Abbreviation of Contrastive Language-Image Pre-training. https://openai.com/blog/clip/, Learning Transferable Visual Models From Natural Language Supervision., A. Radford et al. 2021, https://cdn.openai.com/papers/Learning_Transferable_Visual_Models_From_Natural_Language_Supervision.pdf
[45] An encoder compresses information, and a decoder retrieves the information.
[46] Environment for building customized application that meets business needs provided by Microsoft. It allows for quick development of applications. https://powerapps.microsoft.com/ja-jp/
[47] https://www.infoq.com/jp/news/2021/01/microsoft-license-gpt-3/
[48] Low-code language used for instances in Power Apps. It includes the general-purpose type used for instances in Excel programming, strict type specification, declarative programming and functional programming.
[49] https://news.microsoft.com/ja-jp/2021/05/26/210526-microsoft-introduces-its-first-product-features-powered-by-gpt-3/
[50] PET/iPET: Deep learning training method for natural language processing NLP) model. Abbreviation of Pattern-Exploiting Training, Iterative Pattern-Enhanced Training. It is a training method to achieve the same performance through smaller models.
[51] One of the natural language processing benchmarks. It consists of tests that include diverse tasks, such as question and answer, natural language inference, cross-referencing solution, and resolution of words ambiguity.
[52] ALBERT: A Lite BERT for Self-supervised Learning of Language Representations., Z. Lan et al. 2019, https://arxiv.org/abs/1909.11942

Paper Appendix, Section 1, AI Technology, Chapter 1 9 Creation, for details.
[87] Diffusion Models Beat GANs on Image Synthesis. P. Dhariwal et al. 2021, https://arxiv.org/abs/2105.05233
[88] https://openai.com/dall-e-2/, https://cdn.openai.com/papers/dall-e-2.pdf
[89] MLP: Multi layer perceptron, one of basic structures that is multilayer-coupled to allow for forward transmission of neural networks.
[90] Pay Attention to MLPs H. Liu et al. 2021, https://arxiv.org/abs/2105.08050
[91] SGU: Spatial Gating Unit　Gating mechanism that performs sequence direction calculation of input elements.
[92] An Image is Worth 16x16 Words: Transformers for Image Recognition at Scale, A. Dosovitskiy et al.,ICLR 2021, https://openreview.net/forum?id=YicbFdNTTy
[93] The complete pattern of ocular dominance stripes in the striate cortex and visual field of the macaque monkey. S. LeVay et al. 1985, J. Neuroscience,5
[94] Framelet analysis of some geometrical illusions.2010, H. Arai et al., Japan Journal of Industrial and Applied Mathematics
[95] The Structures of Letters and Symbols throughout Human History Are Selected to Maych Those Found in Objects in Natural Scenes., Changjzi et al. 2006, The American Naturalist
[96] Object Representation in Inferior Temporal Cortex Is Organized Hierarchically in a Mosaic-Like Structure. M. Tanifuji et al. 2013, J. Neuroscience
[97] HyperTransformer: Model Generation for Supervised and Semi-Supervised Few-Shot Learning., A. Zhmoginov et al. 2022, https://arxiv.org/abs/2201.04182
[98] Computer Vision Saizensen Winter2021, 2021 Kyoritsu Shuppan
[99] MetaFormer is Actually What You Need for Vision., W. Yu et al. 2022, CVPR2022, https://arxiv.org/abs/2111.11418
[100] https://ai.facebook.com/research/data2vec-a-general-framework-for-self-supervised-learning-in-speech-vision-and-language
[101] https://imagen.research.google/
[102] https://laion.ai/laion-400-open-dataset/
[103] https://openai.com/blog/instruction-following/
[104] Deep reinforcement learning from human preferences., P.F Christiano et al. 2017, https://proceedings.neurips.cc/paper/2017/file/d5e2c0adad503c91f91df240d0cd4e49-Paper.pdf
[105] Training Compute-Optimal Large Language Models., J. Hoffmann et al. 2022, https://arxiv.org/abs/2203.15556
[106] Flamingo: a Visual Language Model for Few-Shot Learning., J-B. Alayrac et al. 2022, https://arxiv.org/abs/2204.14198
[107] https://ai.googleblog.com/2022/04/pathways-language-model-palm-scaling-to.html, PaLM: Scaling Language Modeling with Pathways., A. Chowdhery et al. 2022, https://arxiv.org/abs/2204.02311
[108] https://www.adept.ai/
[109] https://www.deepmind.com/publications/a-generalist-agent,
[110] Improved protein structure prediction using potentials from deep learning., A. W. Senior et al. 2020, Nature
[111] Highly accurate protein structure prediction with AlphaFold., J. Jumper et al. 2021, Nature
[112] For example, Compositionally restricted attention-based network for materials property predictions., A. Y. Wang et al. 2021, npj Computational Materials, https://www.nature.com/articles/s41524-021-00545-1

40